\documentclass{article}

\usepackage{arxiv}

\usepackage[T1]{fontenc}
\usepackage[utf8]{inputenc}
\usepackage{fancyhdr}

\usepackage[square,comma,numbers,compress]{natbib}
\setcitestyle{nosort}

\usepackage{tikz}
\usetikzlibrary{arrows.meta,positioning,calc,fit,backgrounds,shadows.blur,shapes.geometric}
\usetikzlibrary{arrows.meta,positioning,calc,matrix,decorations.pathreplacing}

\usepackage{microtype}
\usepackage{lmodern}
\usepackage{amsmath,amssymb,mathtools}
\usepackage{amsthm}
\usepackage{makecell}
\usepackage{thmtools}
\usepackage{thm-restate}
\usepackage{bm}
\usepackage{booktabs}
\usepackage{multirow}
\usepackage{siunitx,etoolbox}

\usepackage{booktabs}
\usepackage{graphicx}
\usepackage{float}
\usepackage{xcolor}
\usepackage[normalem]{ulem}
\usepackage{hyperref}
\hypersetup{colorlinks=true,linkcolor=black,citecolor=black,urlcolor=blue}

\newtheorem{remark}{Remark}
\newtheorem{example}{Example}

\title{A Sugeno Integral View of Binarized Neural Network Inference}

\author{
  Ismaïl Baaj \\
  LEMMA, Paris-Panthéon-Assas University \\
  Paris, France \\
  \texttt{ismail.baaj@assas-universite.fr}
  \And
  Henri Prade \\
  IRIT - Toulouse University\\
  31062 Toulouse Cedex 9, France \\
 \texttt{henri.prade@irit.fr}
}
\renewcommand{\shorttitle}{A Sugeno Integral View of Binarized Neural Network Inference}

\hypersetup{
pdftitle={A Sugeno Integral View of Binarized Neural Network Inference},
pdfsubject={cs.AI},
pdfauthor={Ismaïl Baaj, Henri Prade},
pdfkeywords={Sugeno integrals, Binarized neural networks},
}

\begin{document}
\maketitle

\begin{abstract}
In this article, we establish a precise connection between binarized neural networks (BNNs) and Sugeno integrals. The advantage of the Sugeno integral is that it provides a framework for representing the importance of inputs and their interactions, while being equivalent to a set of if–then rules. For a hidden BNN neuron at inference time, we show that the activation threshold test can be written as a Sugeno integral on binary inputs. This yields an explicit set-function representation of each neuron decision, and an associated rule-based representation. We also provide a Sugeno-integral expression for the last-layer score. Finally, we discuss how the same framework can be adapted to support richer input interactions and how it can be extended beyond the binary case induced by binarized neural networks.
\end{abstract}


\section{Introduction}

The development of synergies between Knowledge Representation and Reasoning (KRR) and Machine Learning (ML)
has been emerging as an important line of research \cite{baaj2024synergies,Garcez2019NeuralSymbolicCA,Hitzler2021NeuroSymbolicAI,MARRA2024104062}. It is motivated by the need to combine two complementary abilities: on the one hand,
learning from data for low-level perception, with robustness to noise and ambiguity;
on the other hand, leveraging domain knowledge to reason with explicit rule-based
representations, so as to derive structured conclusions and provide meaningful
justifications.

Several frameworks have been advocated for interfacing learning and reasoning. While the combination of logic and probability remains the dominant option, alternative representations have been proposed
to cope with incomplete or conflicting information and, more generally, to model ignorance explicitly. 
In particular, \cite{denoeux2019logistic} has proposed an evidential view \cite{Shafer1976} of neural-network computations based on belief functions. More recently, explicit links \cite{sum/DuboisP20,fuzzIEEE/BaajPOM21} have also been established between rule-based systems in the setting of possibility theory \cite{DP24} and neural computations, leading to the development of a neuro-symbolic approach \cite{BM25}. 

In this paper, we follow this general direction by establishing links between Sugeno integrals \cite{Sug74} and Binarized Neural Networks (BNN) \cite{hubara2016binarized}. Here, we use \emph{binarized neural network} in the sense of \cite{hubara2016binarized}, where weights and hidden-layer activations are constrained to two values, typically $\{-1,+1\}$. This should be distinguished from the broader notion of binary neural network considered, for instance, in \cite{shi2020tractable}. In binarized neural networks, hidden neurons compute bipolar signed sums over $\{-1,+1\}$ \cite{hubara2016binarized,umuroglu2017finn}, whereas in typical binary neural networks, hidden neurons take $\{0,1\}$ inputs and use step activations, so that each neuron induces a Boolean threshold function \cite{shi2020tractable}. Binarized neural networks have also been widely studied as storage- and computation-efficient models for deep learning on resource-constrained devices, notably in vision tasks such as image classification; see, e.g., \cite{qin2020PR_BNNsurvey,yuan2023comprehensive}.

In decision theory, the Sugeno integral is commonly used as  a qualitative
aggregation function on a finite scale or on the unit interval, and its flexibility comes from the use of a set function, called a ``capacity'', that models
the importance (and possible interaction) of subsets of criteria \cite{grabisch2010decade,grabisch2016set}. Interestingly enough, the Sugeno integral is equivalent to a set of if-then rules \cite{GMS04}. It has also been used in fuzzy rule-based classification \cite{wieczynski2024application}.\\
However, earlier work relating Sugeno integrals and neural networks has mainly used neural networks to learn Sugeno integrals \cite{jia1997using}, rather than deriving an exact Sugeno-integral formulation of neural-network inference. On the other hand, binarized neural networks constrain weights and hidden-layer activations to two values \cite{hubara2016binarized}, which yields threshold computations that are simple and well-structured \cite{umuroglu2017finn}. Existing links between BNNs and knowledge representation have mainly concerned verification issues \cite{narodytska2018verifying}. Our aim is to establish a structural connection between BNNs and Sugeno integrals.

The purpose of this paper is to show that  BNN threshold computations can be expressed exactly in the Sugeno-integral
framework, thereby providing a common representation for qualitative aggregation and binarized neural inference. 
More precisely, we establish the exact correspondence for hidden-layer BNN neurons (Subsection \ref{subsec:hiddenBNNneuronToSugeno}), show that the focal sets of the associated capacity describe minimal activation patterns (Subsection \ref{subsec:hiddenBNNneuronToSugeno}), extend the construction layer by layer across hidden layers (Subsection \ref{subsec:cascadingBNNneuronToSugeno}), and finally show how the last layer can also be represented in this framework, up to an affine rescaling of the resulting $[0,1]$-valued score (Subsection \ref{subsec:lastlayerBNNneuronToSugeno}).

Although
BNN weights and hidden activations take only two values, and our exact Sugeno-integral reformulation inherits this
restriction, the construction clarifies a general correspondence between neuron decision rules and capacities.
Seen through this lens, the BNN threshold rule corresponds to a particularly simple threshold capacity, whereas the
Sugeno framework accommodates much richer set functions modeling importance and interaction among inputs. This perspective
suggests that further connections can be developed in both directions, by using Sugeno-type capacities to model richer
neural computations and, conversely, by interpreting trained network components through the capacities they implicitly
induce.

The paper is organized as follows. Section 2 recalls~the necessary background. Section 3 presents the Sugeno integral representation of BNN neuron computations. A discussion outlines the 
perspectives opened by these connections.

\section{Background}

In the following, we remind the necessary background on binarized neural networks and Sugeno integrals.

\subsection{Binarized Neural Networks}

A binarized neural network (BNN)~\cite{hubara2016binarized} is a neural network in which the weights and the outputs of all hidden layers are constrained to two values, typically $\{-1,+1\}$.
For a hidden neuron at inference time, the computation can be written as a thresholded weighted sum \cite[(Sec.~4.2.2)]{umuroglu2017finn}: for a weight vector $\bar w\in\{-1,+1\}^n$ and a neuron-specific threshold $\theta\in\mathbb{R}$,
\begin{equation}\label{eq:bnn-threshold}
s=\sum_{i=1}^n \bar w_i \bar x_i,
\qquad
\bar y=\begin{cases}
+1,& s \geq \theta\\
-1,& \text{otherwise}
\end{cases},
\end{equation}
where $\bar x\in\{-1,+1\}^n$ is the (binarized) input
 and $\bar y\in\{-1,+1\}$ is the output. 
 In their work introducing BNNs, \cite{hubara2016binarized} model a hidden layer as a linear pre-activation (a weighted sum, optionally with a bias), followed by batch normalization and then a sign activation.
 \cite[(Sec.~4.2.2)]{umuroglu2017finn} show that, at inference time, the composition of batch normalization with the sign activation can be replaced exactly by a comparison of $s=\sum_i \bar w_i \bar x_i$ with a fixed threshold $\theta$ (derived from the batch-normalization parameters), reducing to~\eqref{eq:bnn-threshold}.
Independently, \cite[(Table~1 and Eqs.~(1)--(4))]{narodytska2018verifying} give a representation of BNN layers as a linear weighted sum followed by batch normalization and then binarization, and derive an equivalent threshold form consistent with~\eqref{eq:bnn-threshold}.\\
In~\cite{hubara2016binarized}, all layer inputs are binary except the first layer; the first-layer neuron is computed in a similar form as~\eqref{eq:bnn-threshold} but with an input $u\in\mathbb{R}^n$ (e.g., pixels), i.e., $s=\sum_{i=1}^n w_i u_i$ (optionally $+b$), followed by batch normalization and a sign, which at inference again reduces to thresholding $s$ by a constant, similar to~\eqref{eq:bnn-threshold}.  The last layer commonly produces real-valued class scores (logits) given by a weighted sum (and possibly a bias) before the final decision~\cite{hubara2016binarized,umuroglu2017finn}.

\subsection{Sugeno Integrals}
\label{subsec:sugenointegrals}
Sugeno integrals \cite{Sug74,grabisch2010decade} are a family of qualitative multiple criteria aggregation functions which take  values between the $\min$ and the $\max$ of the aggregated values.
Let $\mathcal{C}$ be a finite set of criteria, and let $\mathcal{L}$  be a scale which is a bounded totally ordered set with a bottom denoted by $0$ and a top denoted by $1$. Moreover $\mathcal{L}$ is supposed to be equipped with an involutive order reversing map denoted by `$1-$'.  A Sugeno integral $S_\mu$ is defined using a {\it capacity} $\mu$ which is a set function from 
$2^{\mathcal{C}}$  to $\mathcal{L}$; $\mu$ is supposed to be increasing in the wide sense ($A \subseteq B \subseteq {\mathcal{C}} \Longrightarrow \mu(A) \leq \mu(B)$), and such that $\mu(\emptyset)=0$ and $  \mu(\mathcal{C})=1$.
Let $x_i \in \mathcal{L}$ denote the evaluation of the item $x$ according to criterion $i \in \mathcal{C}$. The evaluation of $x$ by $S_\mu$ is defined by ($\vee$ and $\wedge$ are the meet and join operations in $\mathcal{L}$) 
\begin{equation}\label{eq:sugenoformulas-vee-wedge}
    S_\mu(x)
= \bigvee_{A\subseteq \mathcal{C}}
\left(\mu(A)\ \wedge\ \bigwedge_{i\in A} x_i\right). 
\end{equation}
Note that when $\mathcal L\subseteq[0,1]$, one often writes $\vee$ and $\wedge$ as $\max$ and $\min$. If $x$ is a binary vector encoding a subset $H\subseteq\mathcal C$, i.e., $x_i=1$ iff $i\in H$, then
$S_\mu(x)=\mu(H)$.
The qualitative M\"obius transform of a capacity $\mu$ is the set function
$\mu_\#:2^{\mathcal C}\to\mathcal L$ defined by (see, e.g.,~\cite{grabisch2004mobius}):
\[
\mu_\#(A)=
\begin{cases}
\mu(A), & \text{if } \mu(A)>\max_{B\subset A}\mu(B),\\
0, & \text{otherwise.}
\end{cases}
\]
A subset $A\subseteq\mathcal C$ such that $\mu_\#(A)>0$ is called a \emph{focal set};
we denote by $\mathcal F^\mu$ the set of focal sets of $\mu$.
Then the Sugeno integral can be written using only focal sets as
$
S_\mu(x)=\bigvee_{A\in\mathcal F^\mu}\Bigl(\mu_\#(A)\ \wedge\ \bigwedge_{i\in A}x_i\Bigr).
$

\cite{GMS04} have pointed out that Sugeno integrals can be represented by a set of   selection rules, or a set of elimination rules having the same threshold for all attributes.
Indeed, any focal set $F$ of a capacity $\mu$ with weight $\mu_\#(F)$ corresponds to the \emph{selection rule}:
$$
\text{If } \forall i\in F,\ x_i \ge \mu_\#(F)\ \text{ then } S_\mu(x)\ge \mu_\#(F).
$$
Moreover, any focal set $A$ of the conjugate capacity of $\mu$ denoted by $\mu^c$ and defined by $\mu^c(B)=1-\mu(\overline{B})$ where $\overline{B}:=\mathcal C\setminus B$, can be associated with the \emph{elimination rule}:
$$
\mbox{If } \forall i \in A,\ x_i \leq 1 -(\mu^c_\#(A))\ \mbox{ then }\ S_\mu(x) \leq 1 -(\mu^c_\#(A)).
$$

\noindent The semantics of Sugeno integrals can also be logically represented by a collection of propositional possibilistic logic formulas of the form $(p_i, \alpha_i)$ \cite{HandbookDP14} where $p_i$ is a propositional formula and $\alpha_i \in [0,1]$, whose semantics is in terms of constraints  $N(p_i)\ge \alpha_i$; $N$ being a necessity measure~\cite{dubois2014logical}. This representation describes the synergies between the criteria, and the importance of satisfying, to a prescribed extent, certain specific combinations thereof.

One can represent a threshold constraint
\begin{equation}\label{eq:thresholdconstraintSugeno22}
    \sum_{i=1}^n w_i x_i \ge \theta
\end{equation}

with $x_i\in\{0,1\}$ and nonnegative weights $w_i\ge 0$ using a $\{0,1\}$-valued capacity and a Sugeno integral.

Let $\mathcal{C}=\{1,\dots,n\}$ and define $\mu:2^{\mathcal{C}}\to\{0,1\}$ by
\begin{equation}\label{eq:SugenoDefthresholdconstraintSugeno22}
\mu(A)=1 \iff \sum_{i\in A} w_i \ge \theta,
\end{equation}
and $\mu(A)=0$ otherwise. Because $w_i\ge 0$, $\mu$ is monotone, hence a capacity.

For $x\in\{0,1\}^n$, let
\[
E(x)=\{i\in\mathcal{C}:x_i=1\}.
\]
Since $x$ is binary,
\[
\bigwedge_{i\in A}x_i=1 \iff A\subseteq E(x),
\]
hence
\[
S_\mu(x)=\bigvee_{A\subseteq E(x)}\mu(A)=\mu(E(x)).
\]
Thus,
\[
S_\mu(x)=1 \iff \sum_{i\in E(x)} w_i \ge \theta  \iff \sum_{i=1}^n w_i x_i \ge \theta.
\]

Moreover, since $S_\mu(x)=S_{\mu_\#}(x)$ and the set of focal elements is
\[
\mathcal{F}^\mu=\{F\subseteq\mathcal{C}:\mu_\#(F)>0\},
\]
one can equivalently write
\[
S_\mu(x)=\bigvee_{F\in\mathcal{F}^\mu}\bigl(\mu_\#(F)\wedge\bigwedge_{i\in F}x_i\bigr).
\]

In this $\{0,1\}$-valued threshold case, any focal set $F\in\mathcal{F}^\mu$ satisfies
\[
\sum_{i\in F} w_i\ge\theta
\qquad\text{and}\qquad
\sum_{i\in F\setminus\{j\}} w_i<\theta \quad \text{for all } j\in F.
\]
Thus, focal elements are minimal combinations of criteria $i$ whose sum of $w_i$'s meets or exceeds the threshold~$\theta$.

\section{Representing BNNs by Sugeno Integrals}
In the following, we show that BNN neuron activations can be computed using Sugeno integrals.
We first show that the BNN neuron activation can be performed in the $\{0,1\}$ binary encoding instead of the $\{-1,1\}$ binary encoding.
We focus on the hidden layers and the last layer, since the first layer typically performs the same thresholded weighted-sum computation as in~\eqref{eq:bnn-threshold} but on non-binary inputs $u\in\mathbb{R}^n$ (e.g., real-valued pixels): a Sugeno-integral formulation for the first layer therefore requires an explicit preprocessing/encoding step, e.g., a normalization mapping the raw input to $[0,1]^n$, or a quantization/encoding mapping it to a binary vector. We treat the last layer separately  because it produces a real-valued score rather than a binary~output.

\subsection{BNN neuron activation in \texorpdfstring{$\{0,1\}$}{0,1} encoding}
\label{subsec:bnnecondingin01}
We first introduce a binary encoding that preserves the neuron decision and allows us to restate the computation with Sugeno integrals.

Fix $\bar w,\bar x\in\{-1,+1\}^n$ and $\theta\in\mathbb{R}$ as in~\eqref{eq:bnn-threshold}.
The goal is to rewrite $s=\sum_{i=1}^n \bar w_i \bar x_i$ and the threshold test $s \geq \theta$, see (\ref{eq:bnn-threshold}), in a $\{0,1\}$ setting. We use a polarized (two-literal) encoding of each coordinate in $\bar w,\bar x$.
Introduce the index set
\[
I:=\{1^+,1^-,\dots,n^+,n^-\}\quad(|I|=2n),
\]
and encode the input $\bar x$ as a binary vector $x\in\{0,1\}^{I}$ by
\begin{equation}\label{eq:xencoding}
x_{i^+}:= \delta^{+1}_{\bar x_i} ,
\qquad
x_{i^-}:= \delta^{-1}_{\bar x_i},
\qquad i\in[n]. \end{equation}
where $[n]=\{1,\dots,n\}$ and we set 
\[ \delta_u^v:= \begin{cases}
    1  &  \text{if} \quad u = v\\
    0  &  \text{if} \quad u \not= v
\end{cases} \quad \text{for all $(u , v)\in\{-1 , + 1\}^2$}.\]
Thus, for each $i$, exactly one of $x_{i^+}$ and $x_{i^-}$ equals $1$.\\
For any $v\in\{0,1\}^{I}$ define its active set  by
\[
E(v):=\{\,j\in I:\ v_j=1\,\}.
\]
Encode the weights by selecting, for each coordinate $i$, the literal that corresponds to the sign of $\bar w_i$:
\begin{equation}\label{eq:Lambda}
    \Lambda(\bar w):=\{\,i^+:\ \bar w_i=+1\,\}\ \cup\ \{\,i^-:\ \bar w_i=-1\,\}\subseteq I.
\end{equation}

\noindent Then $E(x)\cap\Lambda(\bar w)$ collects exactly the coordinates where the input literal $\bar x_i$ agrees with the corresponding weight $\bar w_i$, and hence ($\text{card}$ stands for cardinality):
\[
|E(x)\cap \Lambda(\bar w)|
=\text{card}\{\,i\in[n]:\ \bar x_i=\bar w_i\,\}.
\]
Moreover, since $\bar w_i\bar x_i=+1$ iff $\bar x_i=\bar w_i$ and $\bar w_i\bar x_i=-1$ otherwise, we obtain
\begin{align*}
\sum_{i=1}^n \bar w_i\bar x_i
&= |E(x)\cap \Lambda(\bar w)| - |E(x)\setminus \Lambda(\bar w)| \\
&= |E(x)\cap \Lambda(\bar w)| - \bigl(|E(x)|-|E(x)\cap \Lambda(\bar w)|\bigr) \\
&= 2\,|E(x)\cap \Lambda(\bar w)| - n,
\end{align*}
where $|E(x)|=n$ and $E(x)=(E(x)\cap \Lambda(\bar w))\ \dot\cup\ (E(x)\setminus \Lambda(\bar w))$ (note that $\dot\cup$ denotes disjoint union).

\noindent Therefore, the test $s \geq \theta$, see~(\ref{eq:bnn-threshold}), is equivalent to
$2\,|E(x)\cap \Lambda(\bar w)| - n \geq \theta$, hence
\[
|E(x)\cap \Lambda(\bar w)| \geq \tau,
\qquad \tau:=\left\lceil\frac{\theta+n}{2}\right\rceil,
\]
where $\lceil\cdot\rceil$ denotes the ceiling function. Since $|E(x)\cap \Lambda(\bar w)| \in \{0,1,\dots,n\}$, the inequality $|E(x)\cap \Lambda(\bar w)| \ge \tau$ yields a non-constant decision only when $1 \le \tau \le n$ (otherwise, it is either always true if $\tau \le 0$ or always false if $\tau > n$).
We now express this threshold condition as a Sugeno integral by choosing an appropriate capacity.

\subsection{Computing the hidden BNN neuron output value by a Sugeno integral}
\label{subsec:hiddenBNNneuronToSugeno}

We give an exact Sugeno-integral representation of the hidden-neuron activation in~\eqref{eq:bnn-threshold}.
The integral outputs a value in $\{0,1\}$ and recovers the test $s \geq \theta$ exactly. Another equivalent representation is stated in Remark~\ref{remark:matchbits}.

From $I$, $\Lambda$ and $\tau$ (Subsection \ref{subsec:bnnecondingin01}) and assuming $1\le\tau\le n$,  
define the set function $\mu_{\bar w,\tau}:2^{I}\to\{0,1\}$ by
\[
\mu_{\bar w,\tau}(A)=
\begin{cases}
1, & |A\cap \Lambda(\bar w)| \geq \tau,\\
0, & \text{otherwise},
\end{cases}
\qquad A\subseteq I.
\]
\noindent The semantics of $\mu_{\bar w,\tau}$ is simple, since its focal sets are $\mathcal{F}^{\mu_{\bar w,\tau}} = \{ A\subseteq \Lambda(\bar w) :  |A|=\tau \}$. We study the Sugeno integral:
\[
S_{\mu_{\bar w,\tau}}(x)
:= \bigvee_{A\subseteq I}
\Bigl(\mu_{\bar w,\tau}(A)\ \wedge\ \bigwedge_{j\in A} x_j\Bigr).
\]
Since $x$ is binary (see its definition in \eqref{eq:xencoding}), $\bigwedge_{j\in A}x_j=1$ iff $A\subseteq E(x)$; moreover $\mu_{\bar w,\tau}$ is monotone, hence:
\[
S_{\mu_{\bar w,\tau}}(x)=\bigvee_{A\subseteq E(x)}\mu_{\bar w,\tau}(A)
=\mu_{\bar w,\tau}(E(x)).
\]
Therefore, the following equivalences hold:
\[
S_{\mu_{\bar w,\tau}}(x)=1
 \Longleftrightarrow
|E(x)\cap \Lambda(\bar w)| \geq \tau
\Longleftrightarrow
\sum_{i=1}^n \bar w_i\bar x_i \geq \theta.
\]

Since the Sugeno-integral formulations above output a Boolean value in $\{0,1\}$, we recover the bipolar neuron output $\bar y\in\{-1,+1\}$ in~\eqref{eq:bnn-threshold} by the affine decoding
\[
\bar y \;=\; 2\,S_{\mu_{\bar w,\tau}}(x)\;-\;1.
\]
Equivalently, $\bar y=+1$ iff $S_{\mu_{\bar w,\tau}}(x)=1$, and $\bar y=-1$ otherwise.

We now provide a simple example illustrating this Sugeno-integral representation of a hidden BNN neuron.
\begin{example}\label{ex:hiddenNN}
Consider a neuron with $n=4$, weight vector $\bar w=(+1,+1,-1,+1)$, threshold $\theta=0$, and bipolar input $\bar x=(\bar x_1,\bar x_2,\bar x_3,\bar x_4)\in\{-1,+1\}^4$.
Then
$\tau=\lceil\frac{\theta+n}{2}\rceil=\lceil\frac{0+4}{2}\rceil=2$ and $\Lambda(\bar w)=\{1^+,2^+,3^-,4^+\}$.
The focal sets are all subsets of $\Lambda(\bar w)$ of cardinality $2$. Equivalently, the neuron activates as soon as any two of the following four matched literals are active:
$1^+$, $2^+$, $3^-$, $4^+$.\\
This yields the following minimal rules:
\begin{itemize}
    \item if $\bar x_1=+1$ and $\bar x_2=+1$, then the neuron activates;
    \item if $\bar x_1=+1$ and $\bar x_3=-1$, then the neuron activates;
    \item if $\bar x_1=+1$ and $\bar x_4=+1$, then the neuron activates;
\item if $\bar x_2=+1$ and $\bar x_3=-1$, then the neuron activates;
\item if $\bar x_2=+1$ and $\bar x_4=+1$, then the neuron activates;
\item if $\bar x_3=-1$ and $\bar x_4=+1$, then the neuron activates.
\end{itemize}
Take the input $\bar x=(+1,-1,+1,-1)$. Only position $1$ matches its weight, so the number of matches is $1<2$, and the neuron does not activate. Equivalently, one can compute the weighted sum and check the threshold:
$s=(+1)(+1)+(+1)(-1)+(-1)(+1)+(+1)(-1)=-2<0$.\\
If we flip $\bar x_2$ from $-1$ to $+1$, then the number of matches becomes $2=\tau$, and $s=1+1-1-1=0\geq 0$. The neuron then activates. 
In the Sugeno view, this means that the focal set $\{1^+,2^+\}$ is active, so the activation is explained directly by a minimal sufficient rule. The weighted-sum form reaches the same conclusion, but the Sugeno form makes explicit both which rule is satisfied and one minimal input change that is enough to make the neuron activate.\\
This example shows more clearly how the Sugeno representation provides a direct basis for explanation, whereas this explanatory structure is not made explicit in the weighted-sum view.
\end{example}

In the next subsection, we show that this Sugeno-integral neuron representation can be composed layer by layer, reproducing the network inference computation exactly for the hidden layers.

Below, we also give a match-bit formulation of this Sugeno-integral neuron representation, convenient for analysing a single neuron.
\begin{remark}\label{remark:matchbits}
Fix a neuron with parameters $(\bar w,\theta)$ and let $\tau=\left\lceil\frac{\theta+n}{2}\right\rceil$.
Define the match-bit vector $z\in\{0,1\}^n$ from the polarized input $x$ by
\begin{equation}\label{eq:zencoding}
z_i :=
\begin{cases}
x_{i^+}, & \bar w_i=+1,\\
x_{i^-}, & \bar w_i=-1,
\end{cases}
\qquad i\in[n].
\end{equation}
Equivalently, $z_i=\delta^{\bar w_i}_{\bar x_i}$ and $\sum_{i=1}^n z_i=|E(x)\cap\Lambda(\bar w)|$, hence
$
s \geq \theta \ \Longleftrightarrow\ \sum_{i=1}^n z_i \geq \tau.
$
In particular, $z_i = 1 \iff \bar{x}_i = \bar{w}_i$ and  $z_i = 0 \iff \bar{x}_i = -\bar{w}_i.$\\
Define the support of $z$ by
$
\mathrm{supp}(z):=\{\,i\in[n]: z_i=1\,\}.$ \\
Let $\mu_\tau:2^{[n]}\to\{0,1\}$ be defined by $\mu_\tau(A)=1$ iff $|A|\geq \tau$. Since $z$ is binary,
$
S_{\mu_\tau}(z)
:= \bigvee_{A\subseteq[n]}\Bigl(\mu_\tau(A)\ \wedge\ \bigwedge_{i\in A} z_i\Bigr)
=\mu_\tau(\mathrm{supp}(z)),
$
so $S_{\mu_\tau}(z)=1 \Longleftrightarrow s \geq \theta$.
\end{remark}

\subsection{Cascading hidden layers}
\label{subsec:cascadingBNNneuronToSugeno}
In a multi-layer BNN, computation proceeds layer by layer: the binary outputs of layer $\ell$ are reused as inputs to layer $\ell{+}1$.
Here, ``cascading'' simply means iterating the same construction across layers.
We use the polarized encoding (Subsection~\ref{subsec:bnnecondingin01}) so that the intermediate variables can be defined once per layer.  Let $y^{(\ell)}\in\{0,1\}^{m_\ell}$ denote the vector of Sugeno-integral outputs in hidden layer $\ell$, where
$y^{(\ell)}_j = S_{\mu^{(\ell)}_j}(x^{(\ell-1)})$ for each neuron $j$.
Define the next layer's polarized input variables by, for each $j\in[m_\ell]$,
\[
x^{(\ell)}_{j^+}:=y^{(\ell)}_j,
\qquad
x^{(\ell)}_{j^-}:=1-y^{(\ell)}_j,
\]
so that exactly one of $x^{(\ell)}_{j^+},x^{(\ell)}_{j^-}$ equals $1$.
With this encoding, every neuron in layer $\ell{+}1$ can again be written in the polarized form (with its own $\Lambda(\bar w)$ and threshold $\tau$). Repeating this step for $\ell=1,2,\dots$ reproduces the BNN hidden-layer computation exactly at inference time. In contrast, the match-bit approach (Remark~\ref{remark:matchbits}) is not suitable for a layerwise construction, because the match-bit vector $z$ depends on the weight vector of each neuron in the next layer and therefore cannot be defined once per layer.

\subsection{Computing the last-layer BNN neuron output value by a Sugeno integral}
\label{subsec:lastlayerBNNneuronToSugeno}
In the following, we extend the construction to the last layer, where the network outputs a real-valued score rather than a binary activation.
We treat the  case where the last-layer pre-logit is an affine function of bipolar activations (common in BNN \cite{hubara2016binarized}):
for $\bar x\in\{-1,+1\}^n$, $\bar w^{\mathrm{out}}\in\{-1,+1\}^n$, and $b^{\mathrm{out}}\in\mathbb{R}$, the output score is computed by:
\[
s^{\mathrm{out}}(\bar x)=\sum_{i=1}^n \bar w^{\mathrm{out}}_i\,\bar x_i + b^{\mathrm{out}}.
\]
We use the same polarized encoding as above (Subsection \ref{subsec:bnnecondingin01}). We encode $\bar x$ as $x\in\{0,1\}^{I}$ by
$x_{i^+}:= \delta^{+1}_{\bar x_i}$,
$x_{i^-}:= \delta^{-1}_{\bar x_i}$ and define $E(v):=\{j\in I:\ v_j=1\}$ and
\[
\Lambda(\bar w^{\mathrm{out}}):=\{\,i^+:\ \bar w^{\mathrm{out}}_i=+1\,\}\ \cup\ \{\,i^-:\ \bar w^{\mathrm{out}}_i=-1\,\}\subseteq I.
\]
Then $|E(x)\cap\Lambda(\bar w^{\mathrm{out}})|=\text{card}\{i:\bar x_i=\bar w^{\mathrm{out}}_i\}$ and hence
$
\sum_{i=1}^n \bar w^{\mathrm{out}}_i\,\bar x_i
=2\,|E(x)\cap\Lambda(\bar w^{\mathrm{out}})|-n.
$\\
The Sugeno integral considered is:
\[
S_{\nu_{\bar w^{\mathrm{out}}}}(x)
:= \bigvee_{A\subseteq I}
\Bigl(\nu_{\bar w^{\mathrm{out}}}(A)\ \wedge\ \bigwedge_{j\in A} x_j\Bigr),
\]
\noindent where $\nu_{\bar w^{\mathrm{out}}}(A):=\frac{|A\cap\Lambda(\bar w^{\mathrm{out}})|}{n}.$ For a binary vector $x$, the Sugeno integral reduces to evaluation on the active set:
$
S_{\nu_{\bar w^{\mathrm{out}}}}(x)=\nu_{\bar w^{\mathrm{out}}}(E(x))
=\frac{|E(x)\cap\Lambda(\bar w^{\mathrm{out}})|}{n}\in[0,1].
$
Therefore the last-layer score $s^{\mathrm{out}}(\bar x)$ is recovered from the encoded input $x$ by the affine rescaling
\[
s^{\mathrm{out}}_{\mathrm{enc}}(x)
:=2n\,S_{\nu_{\bar w^{\mathrm{out}}}}(x)-n+b^{\mathrm{out}},
\]
and $s^{\mathrm{out}}_{\mathrm{enc}}(x)\!=\!s^{\mathrm{out}}(\bar x)$ whenever $x$ is the encoding~of~$\bar x$.

\begin{remark}\label{remark:out-matchbits}
An equivalent last-layer formulation uses match bits.
Fix $\bar w^{\mathrm{out}}\in\{-1,+1\}^n$ and define $z\in\{0,1\}^n$ by
$
z_i := \delta^{\bar w^{\mathrm{out}}_i}_{\bar x_i} \text{for } i\in[n],
$
so that $z_i=1$ iff $\bar x_i=\bar w^{\mathrm{out}}_i$.
Then
$$\sum_{i=1}^n \bar w^{\mathrm{out}}_i \bar x_i
= 2\sum_{i=1}^n z_i - n
= 2\,|\mathrm{supp}(z)| - n,
$$
where $\mathrm{supp}(z):=\{\,i\in[n]: z_i=1\,\}$.
Let $\nu:2^{[n]}\to[0,1]$ be the capacity $\nu(A)=\frac{|A|}{n}$.
Since $z$ is binary, the Sugeno integral reduces to
$
S_\nu(z)=\nu(\mathrm{supp}(z))=\frac{|\mathrm{supp}(z)|}{n},
$
and therefore the output score is recovered as
$
s^{\mathrm{out}}(\bar x)=2n\,S_\nu(z)-n+b^{\mathrm{out}}$.
\end{remark}

For a $K$-class output layer, the construction above yields for each class $k\in[K]$ a Sugeno score $S_{\nu_{\bar w^{\mathrm{out}}_k}}(x)\in[0,1]$ and the associated pre-logit $s_k(x):=2n\,S_{\nu_{\bar w^{\mathrm{out}}_k}}(x)-n+b^{\mathrm{out}}_k$. A probability distribution over classes is then obtained by the softmax transformation $P(k\mid x):=\frac{\exp(s_k(x))}{\sum_{j=1}^K \exp(s_j(x))}$. Alternatively, a possibility distribution can be obtained by $\pi_k(x):=\frac{\exp(s_k(x))}{\max_{j=1}^K\exp(s_j(x))}$, which ensures $\max\limits_k \pi_k(x)=~1$.

\section{Discussion}
In the article, we provide an explicit Sugeno integral view of both hidden-layer activations and last-layer scores of binarized neural networks.   This is of interest because the Sugeno integral is a standard qualitative aggregation function, defined
from a capacity \cite{Sug74,dubois2001use,grabisch2010decade,grabisch2016set}. In our setting, the capacity associated
with a hidden neuron has a simple threshold form. Since the inputs are binary and the capacity is monotone, the Sugeno
integral reduces to evaluating the capacity on the set of active inputs, and this recovers the usual threshold test (\ref{eq:bnn-threshold})
used in BNN. This construction is extended for the last layer of BNN.

This view supports simple local explanations, counterfactual reasoning, and  verification;  a concrete illustration is provided in Example~\ref{ex:hiddenNN}.\\
\emph{Explanation.}
For a hidden neuron, the Sugeno representation makes explicit that the neuron outputs $+1$ exactly when at least $\tau$ literals in $\Lambda(\bar w)$ are active in the input, i.e., when $|E(x)\cap\Lambda(\bar w)|\ge \tau$. Equivalently, the neuron outputs $+1$ as soon as there exists a set $A\subseteq \Lambda(\bar w)$ with $|A|=\tau$ such that $A\subseteq E(x)$ (and thus $\mu_{\bar w,\tau}(A)=1$).
Each such focal set gives a \emph{minimal sufficient condition} for activation: a minimal combination of matched literals that, taken together, guarantees that the neuron activates. This rule-based reading is not made explicit in the usual weighted-sum form; see Example~\ref{ex:hiddenNN} for a simple illustration.\\
\emph{Counterfactual reasoning.}
A counterfactual explanation can be described directly on the bipolar input $\bar x$ by flipping a small number of coordinates so as to change the match count $M:=\text{card}\{i\in[n]:\bar x_i=\bar w_i\}=|E(x)\cap\Lambda(\bar w)|$ with respect to the threshold $\tau$. The Sugeno form makes the relevant quantity ($M$ versus $\tau$) and the corresponding minimal input changes directly visible.\\
\emph{Verification.}
Each hidden neuron is described by a monotone threshold capacity with explicit focal sets, and Subsection \ref{subsec:cascadingBNNneuronToSugeno} shows how this description composes across layers via the polarized encoding.\\
For the output layer, $S_{\nu_{\bar w^{\mathrm{out}}}}(x)$ increases when more literals in $\Lambda(\bar w^{\mathrm{out}})$ are active (i.e., belong to $E(x)$), and decreases when fewer are active, so the pre-logit $s^{\mathrm{out}}_{\mathrm{enc}}(x)=2n\,S_{\nu_{\bar w^{\mathrm{out}}}}(x)-n+b^{\mathrm{out}}$ varies accordingly.

While the capacities used here have a simple form and are directly determined by the neuron parameters, which are
binary in a BNN, the Sugeno framework is not limited to this specific form. More general capacities can represent
non-additive importance and interaction between collections of inputs. This observation relates to learning problems
where one would estimate a capacity from input--output samples
\cite{beliakov2020robust,abbaszadeh2020machine,fuzz/baaj2025suglearning}. Moreover, Sugeno integrals are defined on finite ordered scales, which suggests extending this equivalent representation to neural 
networks whose internal values take more than two values. A natural example in this direction is given by quantized neural networks with more than two discrete levels, such as the 8-bit setting studied by~\cite{jacob2018quantization}. Since such quantized values can be viewed abstractly as elements of a finite bounded ordered scale, a link with Sugeno integrals may be possible beyond the binary case.

Furthermore, some authors have suggested that links could be established between Sugeno integrals and spiking neural networks, see ~\cite[end of Section 5]{dubois2019towards} and the French paper \cite{baaj2022reseau}, whose neurons also rely on explicit threshold-based decisions through an accumulation process followed by a threshold test, such as (\ref{eq:thresholdconstraintSugeno22}). The Sugeno-integral representations introduced in the end of Subsection \ref{subsec:sugenointegrals}, see (\ref{eq:SugenoDefthresholdconstraintSugeno22}), through focal sets gathering minimal combinations of activating inputs, may also be relevant in that setting.

Beyond our parallel between BNN threshold activations and Sugeno-integral aggregation, an “integral” viewpoint also appears in recent work aiming to formalize neurosymbolic AI: \cite{DeSmetDeRaedtDefiningNeuroSymbolicAI} define neurosymbolic inference as an integral of the product of a logic function and a belief function over the space of interpretations. They further remark that extensions based on generalised measures (citing Choquet- and Sugeno-type frameworks) are a natural direction for future work.  Our article supports further exploration of Sugeno's integral to define this neuro-symbolic integral.

\bibliographystyle{unsrtnat}
\bibliography{kr-sample}

\end{document}